# Transductive Ordinal Regression

Chun-Wei Seah, Ivor W. Tsang, Yew-Soon Ong

***Abstract*—Ordinal regression is commonly formulated as a multi-class problem with ordinal constraints. The challenge of designing accurate classifiers for ordinal regression generally increases with the number of classes involved, due to the large number of labeled patterns that are needed. The availability of ordinal class labels, however, is often costly to calibrate or difficult to obtain. Unlabeled patterns, on the other hand, often exist in much greater abundance and are freely available. To take benefits from the abundance of unlabeled patterns, we present a novel transductive learning paradigm for ordinal regression in this paper, namely *Transductive Ordinal Regression* (TOR). The key challenge of the present study lies in the precise estimation of both the ordinal class label of the unlabeled data and the decision functions of the ordinal classes, simultaneously. The core elements of the proposed TOR include an objective function that caters to several commonly used loss functions casted in transductive settings, for general ordinal regression. A label swapping scheme that facilitates a strictly monotonic decrease in the objective function value is also introduced. Extensive numerical studies on commonly used benchmark datasets including the real world sentiment prediction problem are then presented to showcase the characteristics and efficacies of the proposed transductive ordinal regression. Further, comparisons to recent state-of-the-art ordinal regression methods demonstrate the introduced transductive learning paradigm for ordinal regression led to the robust and improved performance.**

## I. Introduction

Ordinal regression (OR) is generally defined as the task where some input sample vectors are ranked on an ordinal scale [**?**], [**?**], [**?**]. In a five-star movie rating, for instance, the higher the rating, the better a movie is perceived to be. This rating can be configured as *ordinal class labels* {1,2,3,4,5}, which represents the number of stars a particular movie can be awarded. Hence the class labels are imbued with ordered information, i.e., a sample vector associated with class label 2 has a higher rating (or better) than another having class label 1, and having class label 3 is better off than having class label 1 and 2, and so on. Ordinal regression is also sometimes referred to interchangeably in the literature, as ordinal classification or multi-class classification models [**?**], [**?**], [**?**] with ordered classes. Today, ordinal regression of movie ratings such as the prediction of movie sentiment ratings, represents an important task of the sales personnel as part of their marketing strategy. Besides sentiment prediction, ordinal regression is also used in a wide area of applications that ranges from information retrieval [**?**], [**?**], collaborative filtering [**?**], medical analysis [**?**], gene expression analysis [**?**], to employee selection and prediction of pasture production [**?**].

Chun-Wei Seah, Ivor W. Tsang, Yew-Soon Ong are with School of Computer Engineering, Nanyang Technological University, Singapore 639798, e-mail {Seah0116,IvorTsang,asYSOng}@ntu.edu.sg

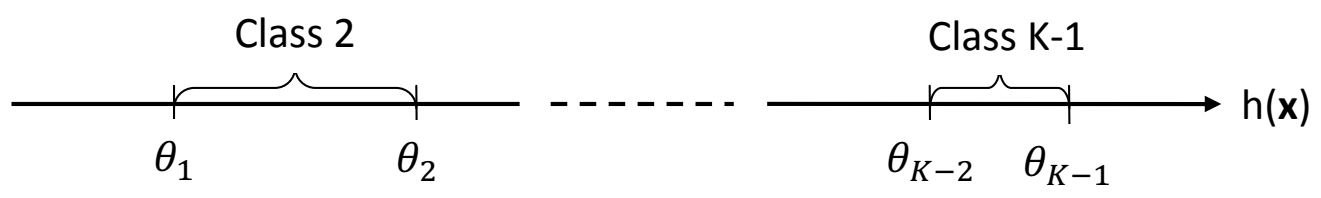

Fig. 1. Threshold model

Initial efforts pertaining to the use of support vector (SV) learning in ordinal regression was reported by Herbrich et al. [**?**]. Their work is based on a threshold model as shown in Fig. 1, in which the threshold values of each ordinal class are estimated. Then, Shashua and Levin [**?**] introduced two approaches for ordinal regression using the large margin principle. The first approach maximizes the margin between adjacent classes, whereas the other maximizes the sum of $K-1$ margins, with $K$ denoting the number of classes.

Both explicit and implicit constraints on the order of the thresholds in the model formulation, referred to as SVOR-EXC and SVOR-IMC in [**?**], [**?**], have also been considered recently. Li and Lin [**?**] extended their work with a framework that transforms the problem of ordinal regression to an extended binary classification, as a generalization of both SVOR-EXC and SVOR-IMC. By deriving the thresholds directly from the support vectors, a more efficient alternative, namely the Reduction Support Vector Machine, was introduced. Last but not least, as opposed to using all $n$ data points, Zhao et al. [**?**] considered $\kappa$ cluster representatives as the training data in SVOR-EXC, leading to significant reduction in the computational complexity, especially for large scale dataset since $\kappa \ll n$.

To summarize, the field of ordinal regression has evolved in the last decade, with a plethora of noteworthy research progress made in supervised learning [**?**], [**?**], [**?**], [**?**], [**?**], [**?**], [**?**], [**?**], [**?**], [**?**]. In spite of the extensive work on this topic, existing methodologies proposed for ordinal regression may be fundamentally bounded by the lack of sufficient class labels found in the data. In particular, it is worth noting that ordinal class labels are often difficult to obtain. Specific tasks such as gene expression [**?**] and cell-phenotype images [**?**] are generally costly to annotate and calibrate due to the need for biological experts. Further, in many realistic applications of science and engineering, it may happen that deriving the labels involves hazardous experiments or the assessment of the label involves extreme conditions in resources [**?**]. A well known example is the movie sentiment problem where ordinal labels of movie ratings are scarce. Moreover, learning all the ordinal boundaries (between pairs of consecutive classes) generally requires considerable amount of labeled data due to the large number of unique class labels involved. Unlabeled data, on the other hand, exists in much greater abundance and are often freely available at zero cost. To take benefits from the abundance of unlabeled patterns, the objective of the present paper is to introduce a novel transductive learning paradigm for ordinal regression, referred to here as *Transductive Ordinal Regression* or *TOR* in short.

TABLE I
*A summary of ordinal regression and related algorithms*

| Learning setting | Algorithm | Type of decision boundaries | Number of Classifiers trained for $K$ ordinal Classes | Cluster Assumption on unlabeled data | Ordinal constraints |
|---|---|---|---|---|---|
| Supervised | SVOR-IMC [?] | Separating two consecutive classes in ordinal regression | 1 classifier with $K-1$ $\theta_k$'s | No | Yes, implicit ordering constraint on $\theta_k$ |
| | SVOR-EXC [?] | Separating two consecutive classes in ordinal regression | 1 classifier with $K-1$ $\theta_k$'s | No | Yes, explicit ordering constraint on $\theta_k$ |
| | RED-SVM [?] | Separating two consecutive classes in ordinal regression | 1 classifier with $K-1$ $\theta_k$'s | No | Yes, $\theta_k$'s are augmented into features |
| Semi-Supervised | TSVM [?] | Separating one class from the rest | $K$ classifiers | Yes | No |
| | TOR (Algo. 1) | Separating two consecutive classes in ordinal regression | 1 classifier with $K-1$ $\theta_k$'s | Yes | Yes, $\theta_k$'s are augmented into features |

The key challenge of TOR design lies in the appropriate incorporation of unlabeled data within the multi-class classification problem formulation with ordinal constraints. This involves the tasks of estimating the ordinal class label of the unlabeled data and the decision function of multiple ordinal classes simultaneously. In TOR, we consider both $p(\mathbf{x})$ and $p(y|\mathbf{x})$. In particular, using the $p(\mathbf{x})$ of both labeled and unlabeled data, we avoid decision boundaries that lie in high density regions (i.e. $p(\mathbf{x})$) [?] by means of cluster assumption [?], [?]. In addition, the extension of classical OR to a Transductive OR paradigm is also non-trivial. To be more precise, current Transductive approaches are not designed to function well on ordinal regression (multi-class[1] with ordering information) problems. Taking this cue, we present in this paper a novel transductive learning paradigm for ordinal regression [?], [?]. In particular, we formulate the ordinal-class problem as an extended binary classification problem, such that the ordinal constraints can be implicitly enforced. Subsequently, a proposed label swapping scheme for multiple class transduction is introduced to derive ordinal decision boundaries that pass through low density region of the augmented labeled and unlabeled data.

A summary of some existing state-of-the-art ordinal regression approaches and the TSVM is outlined in Table I, where the major similarities and differences are explicitly identified with respect to ‘the type of decision boundaries’, ‘the number of classifiers to train for $K$ ordinal classes’ and ‘whether or not cluster assumption and ordinal constraints are imposed’. Notably, TSVM requires $K$ classifiers in order to learn the label of unlabeled data for all $K$ classes at the same time. As such the training process of TSVM is much more time consuming and complex compared to ORs or TOR, since the latter approach only requires single classifier to be trained. Further, the prediction process of TSVM involves $K$ classifiers and does not take the ordinal constraints into considerations. With only a single classifier, the training process of ORs and TOR is clearly more efficient.

For the sake of brevity, the core contributions of the present study are outlined as follows:

1) A transductive learning paradigm of ordinal regression involving labeled and unlabeled data for learning ordinal decision functions is introduced. To the best of our knowledge, the present work serves as the first attempt that addresses the general ordinal regression problem in a transductive setting for a family of commonly used loss functions including hinge loss, logistic loss, Laplacian loss and others listed in Table II.
2) A label swapping scheme for multiple ordinal class transduction is introduced. The proof of strictly monotonic decrease in the objective function is also derived for the swapping scheme. The proposed transductive ordinal regression algorithm is thus established.
3) Numerical study showed that the TOR achieves significant accuracy improvements in terms of mean zero-one and absolute errors when pitted against other state-of-the-art algorithms for ordinal regression and transductive support vector machines.

The rest of this paper is organized as follows: A brief introduction of ordinal regression is provided in Section II. Section III introduces the transductive ordinal regression (TOR) algorithm. Subsection III-A details the initialization of the pseudo-labels for unlabeled data, while the ordinal loss function used in transductive learning by means of label swapping to minimize the structural risk is described in subsection III-B. The parameters that control the importance of the labeled and unlabeled data used in the loss function are then discussed in subsection III-C. Section IV generalizes a family of well established binary functions as potential loss functions in TOR. An instantiation of TOR with hinge loss is also presented in the section. Extensive experimental results on four benchmark datasets and the real-world sentiment prediction problem are reported in Section V. Analysis and discussions pertaining to the experimental results are then provided in Section VI, while the brief conclusions of the present work are drawn in Section VII.

[1]For multi-class without ordering information, readers are referred to [?], [?], [?].

## II. Review of Ordinal Regression

### A. Notation

Throughout the rest of this paper, a superscript $^T$ denotes the transpose of a vector or a matrix. Given $n$ labeled samples: $(\mathbf{x}_i, y_1)$, $(\mathbf{x}_2, y_2)$, ..., $(\mathbf{x}_n, y_n)$ in the data set, where $\mathbf{x}_i \in \Re^p$ represents the $ith$ sample with ordinal class label $y_i \in \{1, 2, ..., K\}$. Consider also a threshold model such as that depicted in Fig. 1, where a $K$ ordinal class problem has $K-1$ ordered thresholds: $\theta_1 < \theta_2 < \ldots < \theta_{K-1}$. Thus, a sample, $\mathbf{x}$, is classified as Class $i$ when the predictive output $h(\mathbf{x}) = \mathbf{w}^T\mathbf{x}$ falls in the range of $\theta_{i-1} < h(\mathbf{x}) \leq \theta_i$, where $\mathbf{w} \in \Re^p$, and $\theta_0 = -\infty$ and $\theta_K = \infty$ are typically assumed. For example, a Class 2 label implies an output that lies between $\theta_1$ and $\theta_2$.

### B. Ordinal Regression as an Extended Binary Classification Model

Ordinal regression using a threshold model generally considers the extended binary classification problem [?] of the form:

$$\begin{aligned} \mathbf{x}_i^k &= (\mathbf{x}_i, \mathbf{e}_k) \in \Re^{p+K-1}, \\ y_i^k &= 1 - 2I[y_i \leq k], \end{aligned} \tag{1}$$

for $k = 1, 2, ..., K-1$. Here $\mathbf{e}_k \in \Re^{K-1}$ denotes a vector with the $kth$ element as value 1 and the rest of the elements having value zero, and $I[\cdot]$ denotes an indicator function that returns 1 if the predicate holds, otherwise a zero is returned. Essentially, each labeled sample $\mathbf{x}_i$ in the original data set is duplicated $K-1$ times, and the $k$th copy is augmented with $\mathbf{e}_k$ and is assigned with a binary label $y_i^k$ in the transformed problem.

A binary classifier with a weight vector

$$\bar{\mathbf{w}} = (\mathbf{w}, -\boldsymbol{\theta}) \in \Re^{p+K-1}, \tag{2}$$

is then learned to predict $y_i^k$ such that $(\mathbf{w}, -\boldsymbol{\theta})^T\mathbf{x}_i^k = \mathbf{w}^T\mathbf{x}_i - \theta_k$. Hence, the threshold $\theta_k$ of the threshold model is estimated using feature augmentation. Subsequently, the predictive ordinal class label of each sample, $\mathbf{x}_i$, is computed as:

$$f(\mathbf{x}_i) = 1 + \sum_{k=1}^{K-1} I[g(\mathbf{x}_i^k) > 0] \tag{3}$$

where $g(\mathbf{x}_i^k) = \bar{\mathbf{w}}^T\mathbf{x}_i^k = (\mathbf{w}, -\boldsymbol{\theta})^T\mathbf{x}_i^k = \mathbf{w}^T\mathbf{x}_i - \theta_k = h(\mathbf{x}_i) - \theta_k$ and $I[\cdot]$ is an indicator function that returns 1 if the predicate holds, otherwise a 0 is returned.

In this manner, besides inheriting the theoretical rigors of binary classifiers, typical caching and optimization techniques such as Sequential Minimal Optimization (SMO) [?], [?] can also be used in ordinal regression.

## III. Transductive Ordinal Regression

In this section, we present the essential components of the proposed TOR algorithm for Ordinal Regression. In particular, we consider the ordinal regression problem where $n$ labeled samples: $(\mathbf{x}_1, y_i), (\mathbf{x}_2, y_2), ..., (\mathbf{x}_n, y_n)$ and $u$ unlabeled samples: $\mathbf{x}_{n+1}, \mathbf{x}_{n+2}, ..., \mathbf{x}_{n+u}$ are available. In what follows, we introduce a novel transductive learning paradigm, referred to here as Transductive Ordinal Regression (TOR), for inferring the labels ($\mathbf{y}^* = \{y_{n+1}, y_{n+2}, ..., y_{n+u}\}$) of $u$ number of unlabeled data instances and modeling the prediction function, $h(\mathbf{x})$, by minimizing the structural risk functional of the form:

**Algorithm 1** Transductive Ordinal Regression (TOR)

Parameters: $C_1$
Inputs: a training set including labeled and unlabeled samples $D_L$=($\mathbf{x}_1$,$y_1$),...,($\mathbf{x}_n$,$y_n$) and $D_U$=$\mathbf{x}_{n+1}$,...,$\mathbf{x}_{n+u}$.
Outputs: predicted labels of $D_U$
// Initialization of unlabeled data's class label
assign $\mathbf{y}^*$ using Algorithm 2
// transductive learning
set $C_2$ =some small value (e.g. $10^{-5}$)
**while** $C_2 < C_1$ **do**
  **repeat**
    $(\mathbf{w}, \boldsymbol{\theta})$:= solve (4) by fixing $\mathbf{y}^*$
    **for** int $k = 1$;$k < K$;$k$++ **do**
      **if** $\exists(i, j)$ satisfying (5) **then**
        if there is more than one $(i, j)$, choose the one with the largest decrease in the loss value
        $y_i = k + 1$
        $y_j = k$
      **end if**
    **end for**
  **until** no label is swapped
  $C_2 = C_2 * 2$
**end while**
**return** $\mathbf{y}^*$

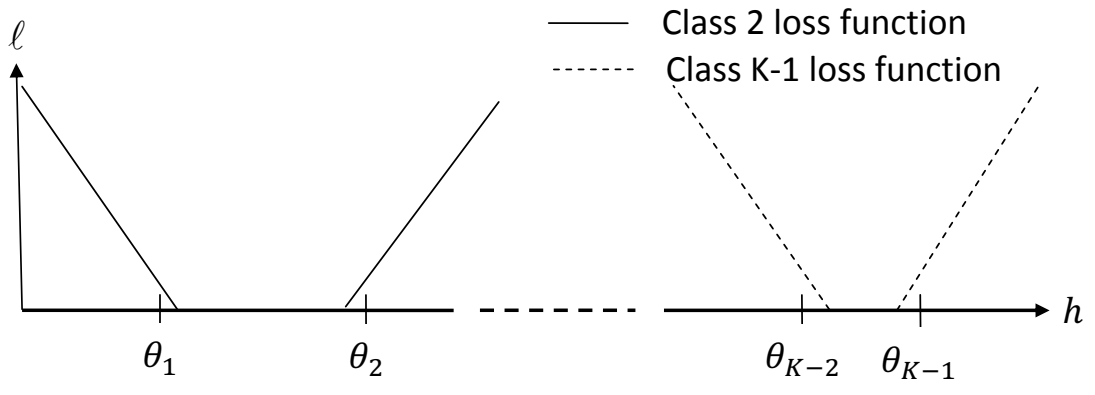

Fig. 2. Loss function for each class in $K$ ordinal class problem

$$\begin{aligned} \min_{h, \boldsymbol{\theta}, \boldsymbol{y}^*} \quad & \tau(h, \boldsymbol{\theta}) + C_1 \sum_{i=1}^{n} \ell_{y_i}(h(\mathbf{x}_i), \boldsymbol{\theta}) \\ & + C_2 \sum_{j=n+1}^{n+u} \ell_{y_j}(h(\mathbf{x}_j), \boldsymbol{\theta}) \\ s.t \quad & \theta_k < \theta_{k+1} \; \forall k \in \{1, ..., K-2\} \end{aligned} \tag{4}$$

where $\tau$ is the regularizer that controls the complexity of $h$ and $\boldsymbol{\theta}$, and $C_1$ and $C_2$ are the parameters that trade-offs the amount of regularization against the loss function $\ell_{y_i}(\cdot)$ on the labeled data and unlabeled data, respectively. Recall that ordinal regression involves a $K$ class problem, hence the loss function in (4) can be represented by $K$ loss functions, where each loss function represents a class depicted in Fig. 2. In another words, each sample, $\mathbf{x}_i$, with a class label, $y_i$, possesses a loss function represented by $\ell_{y_i}(h(\mathbf{x}_i), \boldsymbol{\theta})$.

Through (4), TOR simultaneously learns the order of the decision boundaries, $\boldsymbol{\theta}$, and at the same time the pseudo-labels of unlabeled data with the decision boundaries are enforced

**Algorithm 2** Initialization of pseudo-labels for Unlabeled Data

```
Parameter: $C_1$
Inputs: a training set including labeled and unlabeled samples $D_L$=($\mathbf{x}_1$,$y_1$),...,($\mathbf{x}_n$,$y_n$) and $D_U$=$\mathbf{x}_{n+1}$,...,$\mathbf{x}_{n+u}$
Outputs: $\mathbf{y}^*$ of $D_U$
// Start of algorithm
Count the number of samples $num_k$ in $D_L$ that fall into Class $k$ and then compute $ratio_k = \frac{num_k}{\sum_{i=1}^{K} num_i}$
$(\mathbf{w}, \boldsymbol{\theta})$:=solve (4) with $C_2 = 0$ (i.e. without $D_U$)
Compute the predicted value, $\mathbf{w}^T\mathbf{x}_i$, of $\forall \mathbf{x}_i \in D_U$
Sort $D_U$ in ascending order of the predicted value to form a sorted $D_U^*$
for int $k = 1$;$k$<$K$;$k$++ do
    assign the first $ratio_k$ of unassigned samples in $D_U^*$ with label $k$
end for
assign the rest of unassigned samples in $D_U^*$ as $K$
return $\mathbf{y}^*$
```

to fall on low density regions of both labeled and unlabeled data, while satisfying the cluster assumption. In this manner, majority of the data vectors in the $k$th ordinal class would lie in the range of thresholds, $\theta_{k-1}$ and $\theta_k$, while loss function $\ell_{y_i}(\cdot)$ then caters to the remaining data (*a.k.a*, the outliers) that violates the cluster assumption.

Solving (4) optimally would involve trying out all the possible combinations of assignment for $\mathbf{y}^*$, resulting in a NP hard problem. Hereafter, (4) is solved by first finding $h$ and $\boldsymbol{\theta}$ while fixing $\mathbf{y}^*$, then applying the swapping scheme to update $\mathbf{y}^*$ and repeating the entire process until convergence is reached as outlined in Algorithm 1.

### *A. Pseudo-labels of Unlabeled Data Initialization*

The initialization phase of the TOR focuses on assigning initial pseudo-labels to the unlabeled data. By using a large margin criterion, the optimization problem may lead to trivial solutions, e.g., all unlabeled data are classified with positive labels [**?**], [**?**]. The common practice in transductive learning is to impose some class ratio constraints on the eventual labels of the unlabeled data (*e.g.*, assuming balanced class distribution), where such an assumption has been shown to mitigate the issue of unbalanced output distribution and improves prediction performances [**?**]. Taking this cue, in the TOR, the pseudo-labels of the unlabeled data are constrained to match the class distribution of the labeled data. In particular, the constraints are fulfilled implicitly through the procedure of first training a supervised OR classifier on available labeled data and subsequently sorting the unlabeled data according to the values predicted by the trained supervised OR classifier. Pseudo-labels are then assigned to the sorted set with respect to the class distribution of the labeled data. The procedure to initialize the pseudo-labels of unlabeled data is outlined in Algorithm 2.

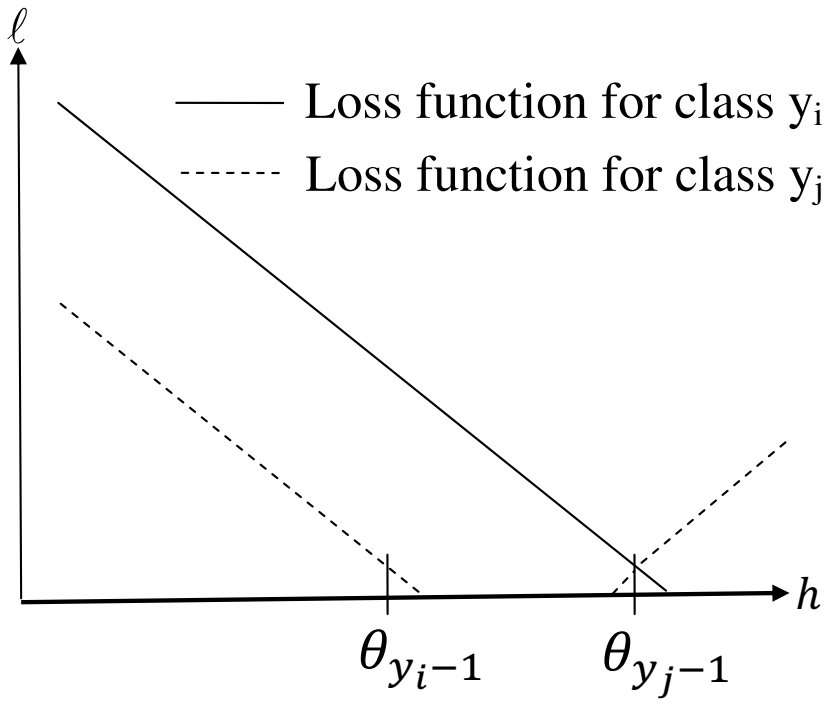

Fig. 3. Two consecutive class loss functions

### *B. Transductive Learning by Label Swapping*

After the initialization phase to define the structural risk functional of (4), the minimization of (4) proceeds with a 2-steps label swapping procedure. The first involves fixing $\mathbf{y}^*$ to solve $h$ and $\boldsymbol{\theta}$. Next, both the derived $h$ and $\boldsymbol{\theta}$ are in-turn fixed to locate suitable $\mathbf{y}^*$ that minimizes objective (4). In what follows, we define the criterion of the ordinal loss function to arrive at solution $\mathbf{y}^*$ that minimizes objective (4).

**Definition 1.** *Loss function $\ell_{y_i}(\cdot)$ is defined with the following properties:*

1) $\forall i, j \quad y_i = y_j - 1,\ h(\mathbf{x}_i) = h(\mathbf{x}_j),\ f(\mathbf{x}_j) < y_j \implies \ell_{y_i}(h(\mathbf{x}_i), \boldsymbol{\theta}) < \ell_{y_j}(h(\mathbf{x}_j), \boldsymbol{\theta})$
2) $\forall i, j \quad y_i = y_j - 1,\ h(\mathbf{x}_i) = h(\mathbf{x}_j),\ f(\mathbf{x}_i) > y_i \implies \ell_{y_i}(h(\mathbf{x}_i), \boldsymbol{\theta}) > \ell_{y_j}(h(\mathbf{x}_j), \boldsymbol{\theta})$

Def. 1 defines the relationship between two consecutive classes. Referring to Fig. 2, a class $k$ loss function is penalized in both directions. For example, the figure depicts a class 2 loss function consisting of a left and a right slanted line. In addition, the relationship between the left section (line) of two consecutive classes is depicted in Fig. 3 (which is a close up version of Fig. 2) and satisfies the first property of Def. 1. In particular, two adjacent class loss functions with the same predicative value, $h$, suggests the lower class loss function exhibits a smaller loss value, $\ell$. In the same manner, the 2nd property of Def. 1 defines the right section of the loss function.

Using the loss function governed by Def. 1, in what follows, we present the details on minimizing the structural risk functional using the proposed label swapping scheme to reduce the loss term in (4). In order to minimize the objective of transductive ordinal regression in (4), the following proposition which extends Theorem 2 in [**?**] from binary class problems to $K$ ordinal class problems, is introduced to cater for the ordinal constraints defined on the unlabeled data.

**Proposition 2.** *For an ordinal loss function defined in Def. 1, swapping the label of two samples $\mathbf{x}_i$ and $\mathbf{x}_j$ from two adjacent classes $y_i$ and $y_j$, i.e., $y_i = y_j - 1$,* (4) *observes a strictly monotonic decrease when $f(\mathbf{x}_i) > y_i$ and $f(\mathbf{x}_j) < y_j$.*

*Proof:* According to Def. 1, the first property assures $\ell_{y_j-1}(h(\mathbf{x}_j), \boldsymbol{\theta}) < \ell_{y_j}(h(\mathbf{x}_j), \boldsymbol{\theta})$ and the second property assures $\ell_{y_i+1}(h(\mathbf{x}_i), \boldsymbol{\theta}) < \ell_{y_i}(h(\mathbf{x}_i), \boldsymbol{\theta})$.

Hence, $\ell_{y_i+1}(h(\mathbf{x}_i),\boldsymbol{\theta}) + \ell_{y_j-1}(h(\mathbf{x}_j),\boldsymbol{\theta}) < \ell_{y_i}(h(\mathbf{x}_i),\boldsymbol{\theta}) + \ell_{y_j}(h(\mathbf{x}_j),\boldsymbol{\theta})$ holds. Through swapping, the last term in (4) will follow a strictly monotonic decrease for fixed $h$ and $\boldsymbol{\theta}$. After the swapping, a new decision function $h'$ and $\boldsymbol{\theta}'$ will be learned for (4). Since (4) is a minimization problem, we have:

$$\tau(h',\boldsymbol{\theta}') + C_1 \sum_{i=1}^{n} \ell_{y_i}(h'(\mathbf{x}_i),\boldsymbol{\theta}') + C_2 \sum_{j=n+1}^{n+u} \ell_{y_j}(h'(\mathbf{x}_j),\boldsymbol{\theta}')$$
$$< \tau(h,\boldsymbol{\theta}) + C_1 \sum_{i=1}^{n} \ell_{y_i}(h(\mathbf{x}_i),\boldsymbol{\theta}) + C_2 \sum_{j=n+1}^{n+u} \ell_{y_j}(h(\mathbf{x}_j),\boldsymbol{\theta}).$$

■

Motivated by Proposition 2 and in the spirit of [?], we propose the swapping of labels between two consecutive classes (i.e. Class $k$ and $k+1$) on unlabeled data for a predictive function $h$ and threshold values $\boldsymbol{\theta}$, when the following conditions have been met:

$$\begin{aligned} \exists i,j \quad & n+1 \le (i,j) \le n+u, y_i = k, y_j = k+1, \\ & f(\mathbf{x}_i) > y_i \ , \ f(\mathbf{x}_j) < y_j \end{aligned} \tag{5}$$

This ensures (4) to strictly decrease upon each swap.

When more than a pair of (*i*,*j*) satisfying the conditions in (5) exists, the pair contributing to the largest decrease in the loss value is selected. Intuitively, this can be viewed as choosing the pair with highest information gain through the strategy.[2]

### C. Control Parameters

$C_1$ and $C_2$ denote the control parameters of the proposed TOR detailed in Algorithm 1. In particular, $C_1$ regulates the tradeoff between mis-classification errors on the labeled samples and the model complexity. In the same way, $C_2$ regulates the tradeoff for the unlabeled samples. $C_1$ denotes a user-specified parameter whereas $C_2$ is heuristically derived in TOR. Typically, $C_2$ is initialized with some small value and gradually increased to approach $C_1$, in the spirit of [?]. This is a common heuristic strategy used to reduce the possibility of premature convergence and getting stuck in poor approximate solution when assigning the labels of the unlabeled data. Note that, when $C_2$ tends to zero, the algorithm becomes a typical supervised learning problem. Therefore, increasing $C_2$ gradually transforms the problem of ordinal regression to TOR. When the stopping criterion pertaining to $C_2$ is reached in TOR, the assigned ordinal class label for the unlabeled data is deemed to converge. Hence, Algorithm 1 serves as a form of heuristic local search for solving (4) by means of approximation.

[2]Note that the training time of this algorithm can be improved by swapping the labels from a set of unique pairs [?] since Proposition 2 guarantees the objective value in (4) to decrease. The study of improving the training time by swapping more than one pair for binary class problems has been shown in [?]. However, premature convergence might result in poor solutions. Hence, there is a tradeoff between the convergence of the training process and the quality of the solution by swapping more than one pairs. For simplicity, swapping only a pair of labels for each adjacent class is considered in the present study.

TABLE II
*A family of binary loss functions can be used in our framework*

| Function | Formulation of loss $\ell_{y_i^k}(a)$ |
|---|---|
| Hinge Loss | $\max\{0, 1-y_i^k(a)\}$ |
| Square Hinge Loss | $(\max\{0, 1-y_i^k(a)\})^2$ |
| Logistic Loss | $\log(1+e^{-y_i^k a})$ |
| Square Loss | $(a-y_i^k)^2$ |
| Laplacian Loss | $\lvert a-y_i^k \rvert$ |

## IV. Generalizing the family of binary loss functions in TOR

In this section, we generalize a family of existing binary functions for use as potential loss function in TOR. In particular, subsection IV-A defines how $K-1$ binary functions can be used as the loss function in TOR. Then, an instantiation of TOR with hinge loss is subsequently showcased in subsection IV-B. Next, label swapping of TOR for $K$ ordinal problem is discussed in subsection IV-C.

### A. Superimposing extended binary functions as the loss function of TOR

Using the representation in the extended binary classification model, binary loss functions that fit in to fulfill the properties of Def. 1 (via superimposing $K-1$ binary loss functions $\ell_{y_i^k}(\cdot)$ defined for each extended binary class $y_i^k \in \{-1,1\}$ of (1)) is as follows:

$$\ell_{y_i}(h(\mathbf{x}_i),\boldsymbol{\theta}) = \sum_{k=1}^{K-1} \ell_{y_i^k}(g(\mathbf{x}_i^k)) \tag{6}$$

where $\mathbf{x}_i^k$ is defined in (1) which incorporates $\theta_k$. Each binary loss function, $\ell_{y_i^k}(\cdot)$, has the following properties:

**Definition 3.** *Binary loss function $\ell_{y_i^k}(\cdot)$ is defined as follows:*

1) $\forall a > 0 \quad \ell_1(-a) > \ell_1(a)$,
2) $\forall i \qquad \ell_{y_i^k}(a) = \ell_{-y_i^k}(-a)$

In Def. 3, the first property defines the binary loss function for $y_i^k = 1$, where higher loss value is assigned to a misclassified sample relative to one that has been correctly inferred. The last property of Def. 3 defines symmetrical positive and negative class loss functions.

**Proposition 4.** *The loss function superimposing $K-1$ binary loss functions that fulfills Def. 3 also fulfills Def. 1*

*Proof:* Let us first prove the first property of Def. 1. We suppose that $y_i = y_j - 1$, $h(\mathbf{x}_i) = h(\mathbf{x}_j)$ and $f(\mathbf{x}_j) < y_j$. From (6), to prove $\ell_{y_i}(h(\mathbf{x}_i),\boldsymbol{\theta}) < \ell_{y_j}(h(\mathbf{x}_j),\boldsymbol{\theta})$ is the same as proving $\sum_{k=1}^{K-1} \ell_{y_i^k}(g(\mathbf{x}_i^k)) - \sum_{k=1}^{K-1} \ell_{y_j^k}(g(\mathbf{x}_j^k)) < 0$. Assume, to the contrary, so

$$\sum_{k=1}^{K-1} \ell_{y_i^k}(g(\mathbf{x}_i^k)) - \sum_{k=1}^{K-1} \ell_{y_j^k}(g(\mathbf{x}_j^k)) \ge 0,$$

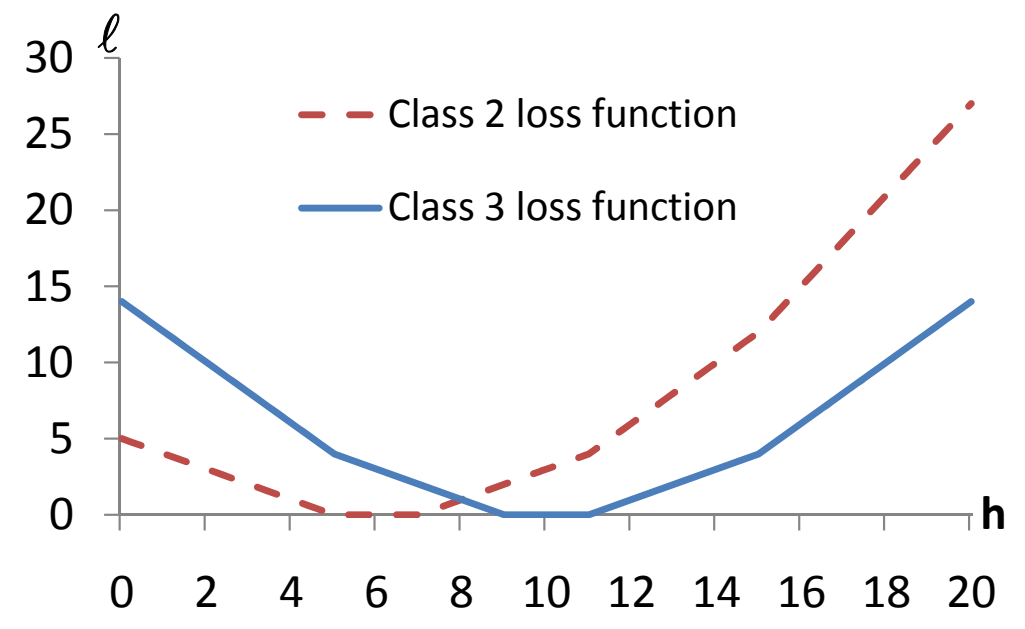

Fig. 4. Loss function $\ell_{y_i}(\cdot)$ using the hinge loss and $K = 5$ with $\theta_1 = 4, \theta_2 = 8, \theta_3 = 12, \theta_4 = 16$.

from (6), we have

$$\begin{aligned}
&\sum_{k=1}^{K-1} \ell_{y_i^k}(g(\mathbf{x}_i^k)) - \sum_{k=1}^{K-1} \ell_{y_j^k}(g(\mathbf{x}_j^k)) \\
=\ & \sum_{k=1}^{y_i-1} \ell_1(g(\mathbf{x}_i^k)) + \sum_{k=y_i}^{K-1} \ell_{-1}(g(\mathbf{x}_i^k)) - \sum_{k=1}^{y_i} \ell_1(g(\mathbf{x}_j^k)) \\
&- \sum_{k=y_i+1}^{K-1} \ell_{-1}(g(\mathbf{x}_j^k)) \\
=\ & \sum_{k=1}^{y_i-1} \ell_1(g(\mathbf{x}_i^k)) + \sum_{k=y_i}^{K-1} \ell_{-1}(g(\mathbf{x}_i^k)) - \sum_{k=1}^{y_i} \ell_1(g(\mathbf{x}_i^k)) \\
&- \sum_{k=y_i+1}^{K-1} \ell_{-1}(g(\mathbf{x}_i^k)) \ \ (\text{since } h(\mathbf{x}_i) = h(\mathbf{x}_j)) \\
=\ & -\ell_1(g(\mathbf{x}_i^{y_i})) + \ell_{-1}(g(\mathbf{x}_i^{y_i})) \\
=\ & -\ell_1(g(\mathbf{x}_i^{y_i})) + \ell_1(-g(\mathbf{x}_i^{y_i})).
\end{aligned}$$

The last equality is derived from the second property of Def. 3. Since $f(\mathbf{x}_j) < y_j$ and $y_i = y_j - 1$, and from (3), we have $\sum_{k=1}^{K-1} I[g(\mathbf{x}_i^k) > 0] < y_i$, which implies $g(\mathbf{x}_i^{y_i}) < 0$, or alternatively $-g(\mathbf{x}_i^{y_i}) > 0 > g(\mathbf{x}_i^{y_i})$. From the first property of Def. 3, we have $\ell_1(-g(\mathbf{x}_i^{y_i}))$ strictly less than $\ell_1(g(\mathbf{x}_i^{y_i}))$. Therefore, $-\ell_1(g(\mathbf{x}_i^{y_i})) + \ell_1(-g(\mathbf{x}_i^{y_i})) < 0$, indicates a contradiction. In the same manner, the second property of Def. 1 can be proven to hold. ■

Therefore, a family of binary loss functions fulfilling the properties in Def. 3 summarized in, but not limited to Table II, can be used to minimize the structural risk functional of TOR framework in (4). The readers are referred to [?], [?] for more details on these loss functions.

### *B. An Instantiation of TOR using Hinge loss*

As mentioned in Section IV-A, our proposed framework can cater to several commonly used loss functions that satisfies Def. 3 to minimize the structural risk functional of (4). Here, we illustrate an instantiation of TOR based on the hinge loss, since it is commonly used in SVM and satisfies Def. 3. For a particular labeled data, $\{\mathbf{x}_i, y_i\}$, and using the extended binary classification model representation with the bias term included in the decision function, the extended binary loss function $\ell_{y_i^k}(\cdot)$ for a particular threshold $\theta_k$ can be derived as:

$$\max\{0, 1 - y_i^k(\bar{\mathbf{w}}^T \mathbf{x}_i^k - b)\} \quad (7)$$

where both the $\theta_k$ augmented $\mathbf{x}_i^k$ and $\bar{\mathbf{w}}^T$ are defined in (1) and (2), respectively.

From (7), the ordinal loss function $\ell_{y_i}(\cdot)$ superimposing the $K-1$ parts satisfies Def. 1 and becomes:

$$\sum_{k=1}^{K-1} \max\{0, 1 - y_i^k(\bar{\mathbf{w}}^T \mathbf{x}_i^k - b)\} \quad (8)$$

as depicted in Fig. 4.

Let $\tau(h, \boldsymbol{\theta}) = \frac{1}{2}\|\bar{\mathbf{w}}\|^2 = \frac{1}{2}\|\mathbf{w}\|^2 + \frac{1}{2}\|\boldsymbol{\theta}\|^2$ (as derived from (2)) and the ordinal loss function $\ell_{y_i}(\cdot)$ as (8), then considering the structural risk of labeled data in (4), the extended binary classification formulation for ordinal regression [?] can be derived as:

$$\begin{aligned}
\min_{\mathbf{w}, b, \boldsymbol{\theta}, \xi_i^k} \quad & \frac{1}{2}\|\mathbf{w}\|^2 + \frac{1}{2}\|\boldsymbol{\theta}\|^2 + C_1 \sum_{i=1}^{n} \sum_{k=1}^{K-1} \xi_i^k, \\
s.t. \quad & y_i^k(\mathbf{w}^T \phi(\mathbf{x}_i) - \theta_k - b) \geq 1 - \xi_i^k, \\
& \xi_i^k \geq 0, \ \ \forall i \in \{1, ..., n\}, k \in \{1, ..., K-1\},
\end{aligned} \quad (9)$$

where $\phi : \Re^p \mapsto \mathcal{F}$ denotes the nonlinear feature mapping induced by a kernel function, and $\mathbf{w}$ is also in $\mathcal{F}$. Thus, the decision functions in (9) become nonlinear by virtue of the *kernel trick* [?]. $\xi_i^k$ denotes the slack variable that caters for the error committed by $\mathbf{x}_i$ at the $k$th decision boundary.

With transductive learning, the labels of the unlabeled data in (4) through (9) are then optimized by:

$$\begin{aligned}
\min_{\mathbf{y}, \mathbf{w}, b, \boldsymbol{\theta}, \xi_i^k} \quad & \frac{1}{2}\|\mathbf{w}\|^2 + \frac{1}{2}\|\boldsymbol{\theta}\|^2 + C_1 \sum_{i=1}^{n} \sum_{k=1}^{K-1} \xi_i^k \\
& + C_2 \sum_{j=n+1}^{n+u} \sum_{k=1}^{K-1} \xi_j^k, \\
s.t. \quad & y_i^k(\mathbf{w}^T \phi(\mathbf{x}_i) - \theta_k - b) \geq 1 - \xi_i^k, \\
& \xi_i^k \geq 0, \ \forall i \in \{1, ..., n\}, k \in \{1, ..., K-1\}, \\
& y_j^k(\mathbf{w}^T \phi(\mathbf{x}_j) - \theta_k - b) \geq 1 - \xi_j^k, \\
& \xi_j^k \geq 0, \ \forall j \in \{n+1, ..., n+u\}, \\
& \qquad k \in \{1, ..., K-1\}.
\end{aligned} \quad (10)$$

Note that the ordered constraints on the thresholds in (4) are implicitly fulfilled in (9) and (10) (see the proof in [?]). Recall that, $\{y_{n+1}, y_{n+2}, ..., y_{n+u}\}$ is denoted by $\mathbf{y}^*$. For a fixed $\mathbf{y}^*$, the dual form of the inner minimization problem in (10) then becomes:

$$\begin{aligned}
\max_{\boldsymbol{\alpha}} \quad & \sum_{i=1}^{n+u} \sum_{k=1}^{K-1} \alpha_i^k \\
& - \frac{1}{2} \sum_{i=1}^{n+u} \sum_{j=1}^{n+u} \sum_{k=1}^{K-1} \sum_{k'=1}^{K-1} \alpha_i^k \alpha_j^{k'} y_i^k y_j^{k'} \kappa(\mathbf{x}_i^k, \mathbf{x}_j^{k'}) \\
s.t. \quad & 0 \leq \alpha_i^k \leq C_1, \ \forall i \in \{1, ..., n\}, k \in \{1, ..., K-1\} \\
& 0 \leq \alpha_j^k \leq C_2, \ \forall j \in \{n+1, ..., n+u\}, \\
& \qquad k \in \{1, ..., K-1\}, \\
& \textstyle\sum_{i=1}^{n+u} \sum_{k=1}^{K} \alpha_i^k y_i^k = 0,
\end{aligned} \quad (11)$$

where $\kappa(\mathbf{x}_i^k, \mathbf{x}_j^k{}') = \phi(\mathbf{x}_i)^T \phi(\mathbf{x}_j') + \mathbf{e}_k^T \mathbf{e}_{k'}$ is the resultant kernel evaluation of $\mathbf{x}_i^k$ and $\mathbf{x}_j^k{}'$, and $\alpha_i^k$ is the Lagrangian multiplier for the inequality constraint in (10). Note this dual is in the form of a quadratic programming (QP) problem, and thus can be easily solved using standard SVM solvers.

In Algorithm 1, one can use (10) to solve (4) while fixing $\mathbf{y}^*$ and then apply the swapping scheme (5) to update $\mathbf{y}^*$. The entire process is then repeated until convergence is reached.

### C. Discussion of label swapping for K ordinal class problem

The proposition 2 for TOR is a generalization of $K$ ordinal class problem, hence the proposition also applies to the binary class problems described in [**?**]. However, the TSVM in [**?**] cannot handle ordinal classification problems elegantly. For example, a data $\{\mathbf{x}, y = 3\}$ in a 5 class problem can be augmented to form binary data using (1) as $\{(\mathbf{x}, e_1), 1\}, \{(\mathbf{x}, e_2), 1\}, \{(\mathbf{x}, e_3), -1\}, \{(\mathbf{x}, e_4), -1\}$. However, swapping with another data vector may cause the dataset to violate the ordinal properties defined in (1) (e.g., $\{(\mathbf{x}, e_1), -1\}, \{(\mathbf{x}, e_2), 1\}, \{(\mathbf{x}, e_3), -1\}, \{(\mathbf{x}, e_4), -1\}$). In contrast, proposition 4 proved that TOR addresses this elegantly by generalizing the ordinal loss function to include commonly used binary loss functions.

## V. Experiments

In this section, we investigate the efficacy of several state-of-the-art ordinal regression algorithms and the proposed transductive ordinal regression, which are described in Table I, on a set of benchmark datasets and the task of sentiment prediction. Since existing ordinal regression models can deal with labeled data only, comparison to three ordinal state-of-the-art algorithms trained with labeled data, are also considered in the present study (namely RED-SVM[3] using (9), SVOR-EXC[4] and SVOR-IMC[4]).

To investigate the effect of cluster assumption on the unlabeled data, comparison to the Multi-class transductive SVM (M-TSVM) [**?**] is also considered by using a multi-class training paradigm. In the experimental study, the M-TSVM is trained using both labeled and unlabeled data based on a one-versus-rest approach. Since the performance of M-TSVM is very sensitive to the balance constraints on the labels of the unlabeled data, a strategy similar to that proposed in Section III-A, i.e., taking the class ratio, $ratio_k$, from the labeled data, as the balance constraints imposed on the labels of the unlabeled data, is also considered for M-TSVM. Taking the $k$th class for example, the constraint enforces the proportion of Class $k$ to the rest of the unlabeled data as $ratio_k : 1 - ratio_k$. With the inclusion of M-TSVM, the impacts of ordinal knowledge on the performance metrics can be analyzed.

### A. Experimental Setup

For each data set, the labeled data are randomly split into different sizes (100, 150, 200, 250, 300, 350 and 400). Let $s$ denotes the sample size of each dataset described in Tables III and IV, $s - 400$ samples then form the set of unlabeled data.

The cost parameter $C_1$ of each algorithm is determined using a five-fold cross-validation procedure with $log_{10}C_1 \in \{-3, -2, -1, 0, 1, 2, 3, 4, 5\}$. To report statistically significant results on the unlabeled data, the average test performances of 20 independent realizations are presented.

To measure the classification error of the samples, **mean zero-one error** is employed as the performance metric and is defined as:

$$\frac{1}{u} \sum_{i=n+1}^{n+u} I[y_i^* \neq y_i^t] \tag{12}$$

where $I[\cdot]$ denotes an indicator function that returns 1 if the predicate holds, otherwise a 0 is returned, and $y_i^*$ and $y_i^t$ are the predicted label of the respective algorithm and the true class label, respectively.

To measure how far the predicted class label of the samples differ from their true class label, the **mean absolute error** is employed here as the performance metric, which is defined as:

$$\frac{1}{u} \sum_{i=n+1}^{n+u} |y_i^* - y_i^t| \tag{13}$$

where $|\cdot|$ denotes the absolute operation.

### B. Benchmark data sets

TABLE III
*Benchmark datasets for ordinal regression*

| Dataset | Sample Size | # Features |
|---|---|---|
| Abalone | 4,177 | 8 |
| Bank | 8,192 | 32 |
| California | 20,640 | 8 |
| Census | 22,784 | 16 |

Four commonly used benchmark datasets[5] (Abalone, Bank, California and Census) in ordinal regression problems are considered in the present study. The statistics of these benchmark datasets are summarized in Table III. These datasets were preprocessed with a quantization level of $K = 5$. For all algorithms, we considered the perceptron kernel [**?**], which is defined as follows:

$$\Delta_p - ||\mathbf{x} - \mathbf{x}^{'}||_2,$$

where $\Delta_p$ denotes a constant. As discussed in [**?**], perceptron kernel can be used by SVM to construct infinite ensemble of classifiers over perceptrons. In other words, the resultant SVM classifier using perceptron kernel is equivalent to a neural network with one hidden layer containing infinite hidden neurons. Moreover, based on the Karush Kuhn Tucker (KKT) conditions, $\sum_{i=1}^{n+u} \sum_{k=1}^{K-1} \alpha_i^k y_i^k = 0$ as derived from (11), the term $\Delta_p$ can be set to zero without changing the objective value of the dual SVM formulation [**?**]. As such, here we

[3]http://www.work.caltech.edu/~htlin/program/libsvm/#ordinal
[4]http://www.gatsby.ucl.ac.uk/~chuwei/svor.htm
[5]http://www.liaad.up.pt/~ltorgo/Regression/DataSets.html

consider the simplified perceptron kernel with $\Delta_p = 0$ in the experimental study[6]

TABLE IV
*Data sets for sentiment prediction*

| Dataset | Sample Size | # Features |
|---|---|---|
| Book | 5,501 | 17,862 |
| DVDs | 5,118 | 19,059 |
| Electronics | 5,901 | 10,728 |
| Kitchen Appliances | 5,149 | 9,230 |

### C. Synthetic data set

A synthetic data set with various degrees of cluster assumption is created based on our generator described in Algorithm 3 to study the performances of transductive TOR versus non-transduction RED-SVM.

**Algorithm 3** Synthetic Data Set Generator

1: Inputs: $y \in [1, ..K]$, where $K$ is the number of ordinal classes, $p$ is a parameter to control the strength of cluster assumption
2: **for** int $d = 1$; $d \leq 2000(K + 2)$; $d$++ **do**
3: **if** $d \in [2000(y - 1), 2000(y + 2)]$ **then**
4: **if** rand()$< 0.01$ **then**
5: $x^d$ =rand()
6: **else**
7: $x^d = 0$
8: **end if**
9: **else**
10: **if** rand()$< 0.01p$ **then**
11: $x^d$ =rand()
12: **else**
13: $x^d = 0$
14: **end if**
15: **end if**
16: **end for**
17: **return** $\mathbf{x}$

Recall that the cluster assumption holds when each class is more separable by a particular set of features, hence line 3 in Algorithm 3 defines the set of features $S_y$ belonging to a particular class $y$. Specifically, a rand() function is used to generate a number $x^d$, which is randomly drawn from a uniform distribution in the interval of 0 and 1. To simulate input vectors with $< 0.01$ probability of sparse features for $x^d \in S_y$, we define $x^d = rand()$, otherwise $x^d = 0$. To define the degree of cluster assumption on feature $x^d \notin S_y$, we introduce parameter $p$ and assign feature $x^d$ with some random at probability of $0.01p$; otherwise, $x^d = 0$. Note, a higher $p$ value lead to greater overlapping among classes, thus a lower degree of cluster assumption. In the experiment, we consider $p = (0.0, 0.1, ..., 0.9)$ and $K = 5$. We randomly generate 20 sets of 2500 examples, and use 200 examples as the labeled data while the remaining as unlabeled data. In addition, the data is normalized as $\frac{\mathbf{x}}{||\mathbf{x}||}$, and with linear kernel used in the experimental study.

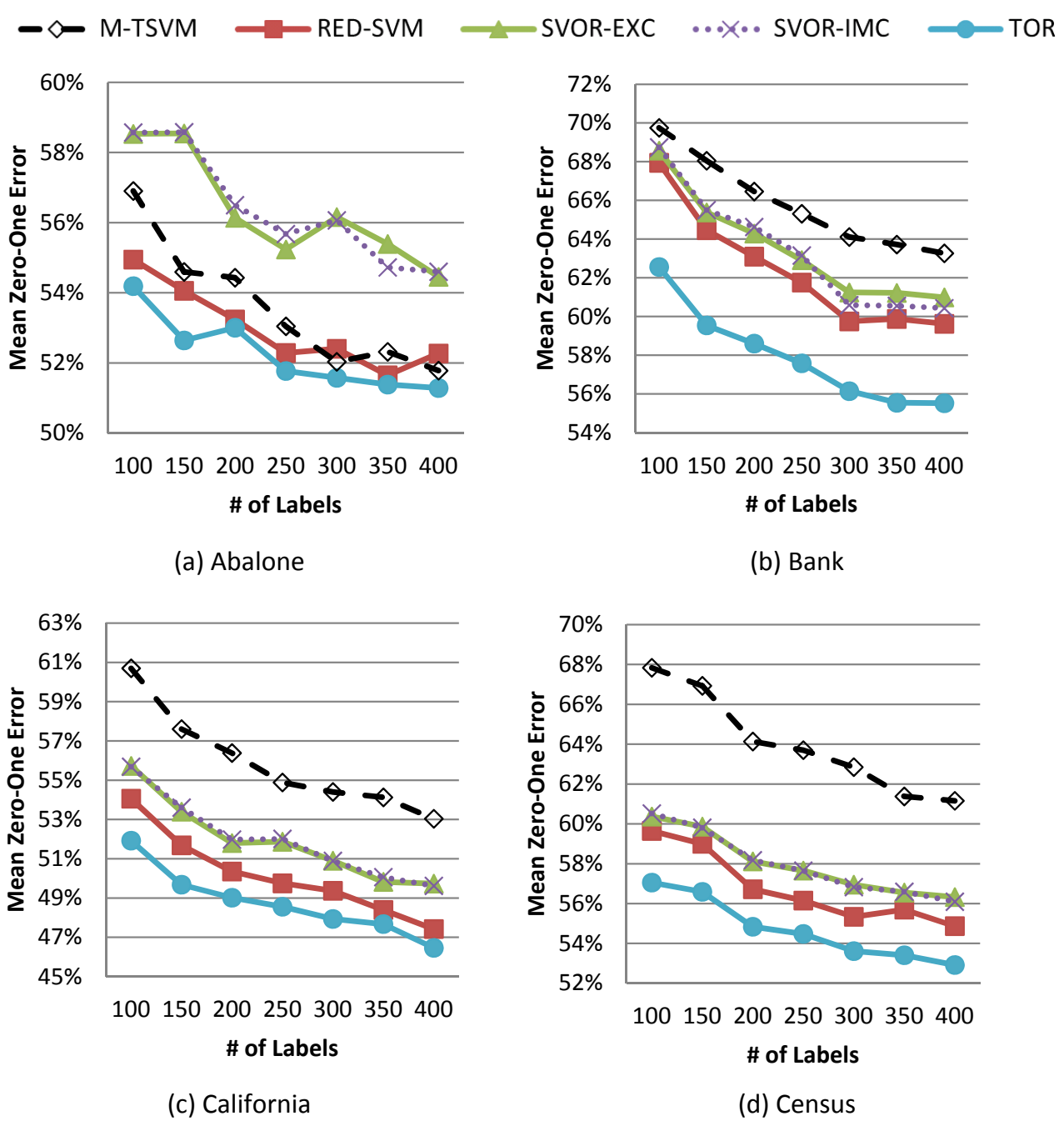

Fig. 5. Mean Zero-One Error on benchmark datasets

### D. Sentiment data sets

The task of sentiment prediction is to predict the star rating of each review. The datasets for sentiment prediction[7] as defined [**?**] were generated from *Amazon.com*, and comprise four categories of product reviews: *Book*, *DVDs*, *Electronics* and *Kitchen appliances*. The reviews consist of five ordinal rating label ranging from 1 to 5. A higher rating means a better review feedback. The details pertaining to the sample and feature size of the sentiment datasets are summarized in Table IV.

In the experimental study, we further preprocessed the datasets by removing all stop-words, normalizing each feature and performing stemming. Finally, each feature of a review is represented by its respective *tf-idf* value. The inner product of two reviews is defined using the cosine similarity, with linear kernel used in the experiments.

[6] Perceptron kernel was reported to offer competitive results to Gaussian Kernel [**?**], but a benefit of perceptron kernel lies in the higher computational efficiency, which has been shown to be more than 10 times faster than Gaussian Kernel. Furthermore, perceptron kernel does not have any additional kernel parameter to be configured. In some previous study on ordinal regression problems [**?**], [**?**], the perceptron kernel was also reported to attain higher accuracies than using Gaussian Kernel.

[7] www.cs.jhu.edu/~mdredze/datasets/sentiment/

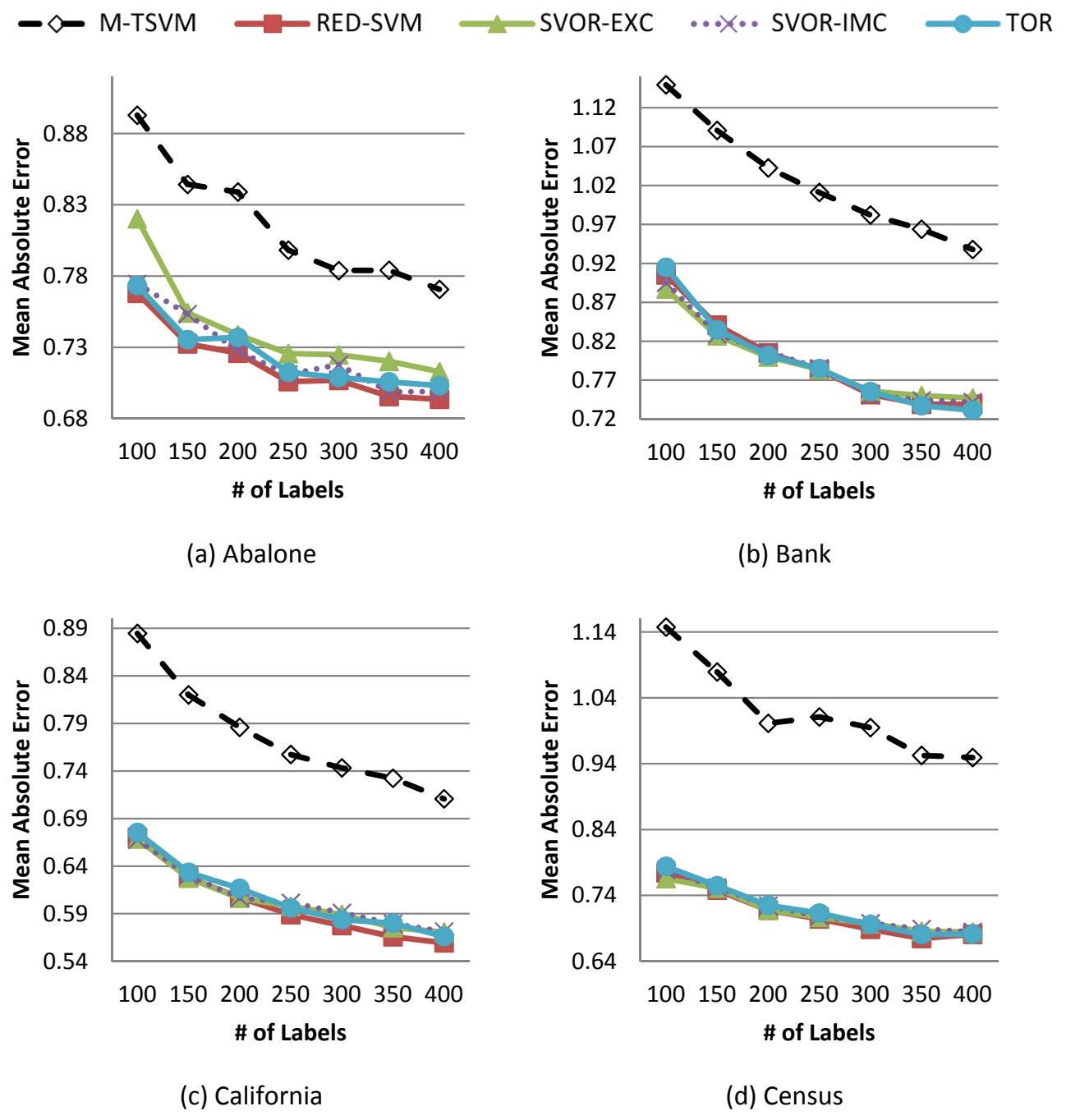

Fig. 6. Mean Absolute Error on benchmark datasets

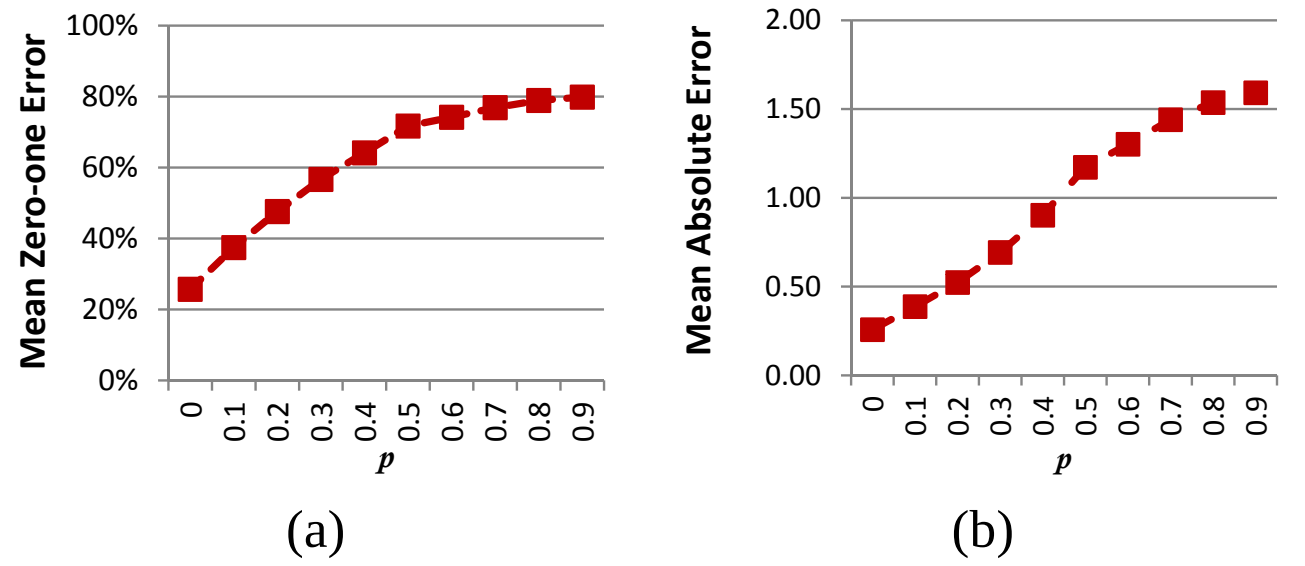

Fig. 7. Analysis of RED-SVM using the class distribution of the labeled data for classification (i.e., the label initialization phase of TOR), on the dataset with various strengths of cluster assumption. SubFigs. (a) and (b) depict the mean zero-one and mean absolute errors, respectively. A higher $p$ value weakens the cluster assumption.

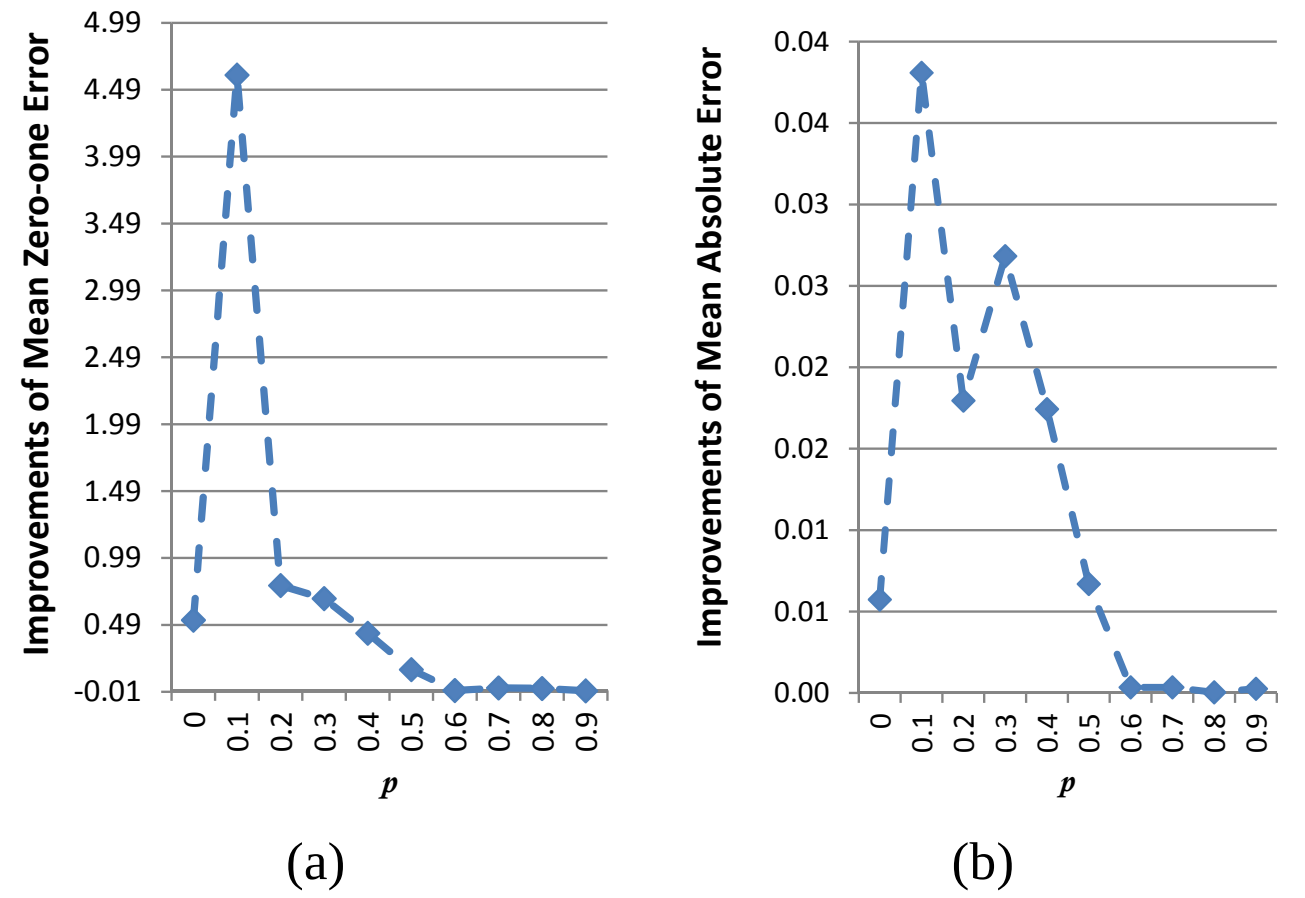

Fig. 8. Analysis of TOR, on the dataset with various strengths of cluster assumption, after the label initialization phase (i.e., RED-SVM using the class distribution of the labeled data for classification). SubFigs. (a) and (b) depict the differences (improvements) of mean zero-one and mean absolute errors, respectively, between TOR reaches convergence in Algorithm 1 and after TOR initializes the labels. A higher $p$ value weakens the cluster assumption.

## VI. Discussions on Experimental Results

### A. Results on Benchmark and synthetic Datasets

On the benchmark and synthetic datasets, we performed experiments for $K = 5$ to assess the predictive performance of various state-of-the-art algorithms. The experimental results of benchmark and synthetic datasets are discussed in subsections VI-A1 and VI-A2, respectively.

*1) Mean Zero-One and Absolute Errors on Benchmark Dataset:* The results of mean zero-one error for each benchmark dataset are summarized in Fig. 5. As observed from the figures, both SVOR-IMC and SVOR-EXC exhibit similar results on all the datasets considered. RED-SVM on the other hand manifests significant improved performances over SVOR-IMC and SVOR-EXC on all the datasets, which is in line with that obtained in [**?**]. Notably, the proposed transductive ordinal regression algorithm, TOR, exhibits the best performances across all experiments. As shown in Fig. 5, TOR reports a minimum of 2% and up to 6% improvements, relative to SVOR-IMC and SVOR-EXC.

As discussed in [**?**], the data in high dimensional feature space such as text documents and sentiment data usually follows the cluster assumption. From the Table III and IV, the number of features of the Bank, Census and Sentiment data sets are higher. From the results reported in Fig. 5, we observed that the improvements of performance of TOR over RED-SVM are higher on the Bank and Census. This is possibly due to the Bank and Census having higher feature dimension so the datasets satisfy the cluster assumption better.

On the manifest of transductive learning, M-TSVM displays the worst performance on most of the experiments, relative to the other algorithms considered, especially on the California and Census datasets in Fig. 5. This is unsurprising since M-TSVM is designed to deal with multi-class problems that does not make use of ordinal information available in the data. Without the use of ordinal knowledge, transduction to infer the correct label of unlabeled data becomes ever more challenging.

Next, we analyze the mean absolute errors of the benchmark regression dataset depicted in Fig. 6. The results indicate that M-TSVM, which does not impose any ordinal constraints, performed badly on all the datasets, as observed in the subfigures. On the other hand, algorithms that use the ordinal information are noted to attain competitive mean absolute errors. While emerging as superior in mean zero-one error, TOR did not top in terms of mean absolute error. We hypothesize this is due to the datasets containing continuous response variables, i.e., regression problems that have been manually quantized into 5 ranks. In Section VI-A2, we will validate our hypothesis on a synthetic dataset.

*2) Mean Zero-One and Absolute Errors on a Synthetic Ordinal Regression Dataset:* Here, we analyze the label swapping procedure of the transductive approach, i.e., TOR, after the non-transductive approach, i.e., RED-SVM, using the class distribution of the labeled data for classification (the label initialization phase of TOR). The results summarized in Figure 7 indicate that the mean zero-one and absolute errors of non-transductive RED-SVM approach deteriorates with decreasing degrees of cluster assumptions (*i.e.*, configured via increasing parameter $p$). Similarly, the proposed transductive approach, i.e., TOR, which leverages the cluster assumption of the unlabeled data, exhibits lower improvements in mean zero-one and absolute errors when the degree of cluster assumption decreases (*i.e.*, $p \geq 0.2$), as depicted in Figure 8. On the other extreme, when the cluster assumption holds strong (*i.e.*, $p = 0$), the improvements in both mean zero-one and absolute errors are observed to be smaller than that for $p = 0.1$. This can be reasoned by the decision boundaries of RED-SVM lying in the low density regions of the labeled and unlabeled data when the cluster assumption holds strong. Finally, when the cluster assumption does not hold (*i.e.*, $p \geq 0.6$), both transductive and non-transductive approaches fail.

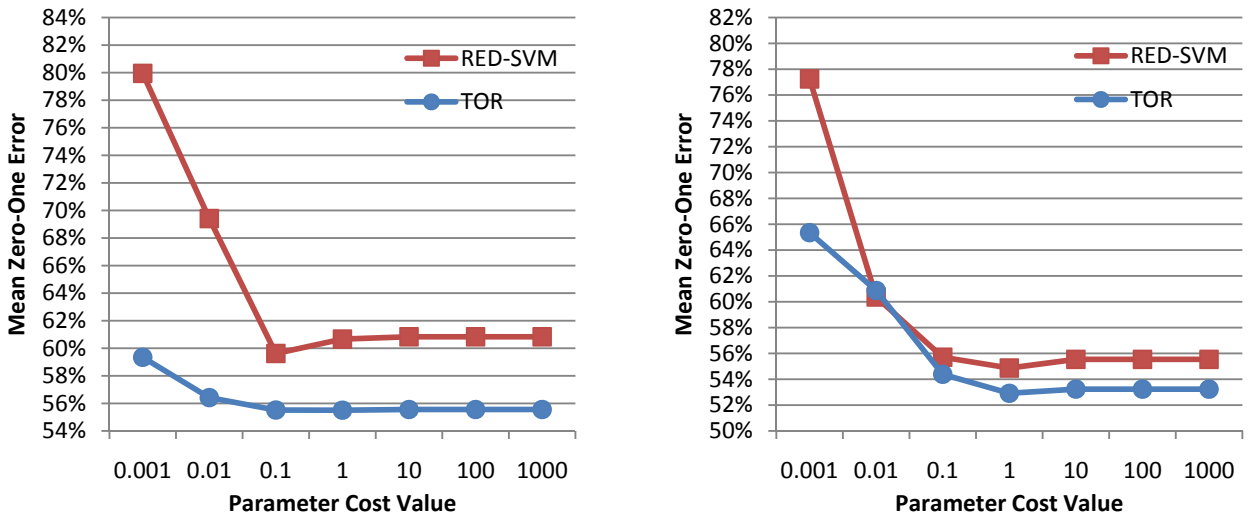

(a) Bank (b) Census

Fig. 9. Mean Zero-One Error varies different $C_1$ values

Later in Section VI-B, our experimental study shows that TOR attains significantly larger improvements over RED-SVM in both mean zero-one and absolute errors on the real world sentiment datasets than on the benchmark datasets. The reason being that, similar to the synthetic data, the real world sentiment datasets are composed of sample data which lie in sparse high dimensional feature space, so the datasets satisfy the cluster assumption more rigorously than the benchmark datasets, since the latter contain continuous response variables that have been artificially quantized to form the ordinal labels.

*3) Sensitivity of $C_1$ Parameter:* In this subsection, we investigated the sensitivity of RED-SVM and TOR methods for different $C_1$ parametric configurations, particularly in the discrete steps of $log_{10}C_1 \in \{-3, -2, -1, 0, 1, 2, 3\}$. We performed the experiments for $K = 5$ and with 400 labeled data. The results depicted in SubFigs. 9 (a) and (b) for Bank and Census datasets, respectively, denote the average test performances of 20 independent realizations. TOR is observed to achieve improved performance on all the settings considered, and exhibit a more stable mean zero-one error than RED-SVM across the range of $C_1$ values. The performance of RED-SVM, on the other hand, is noted to be highly sensitive to the changes in $C_1$ values. The robustness in TOR can be attributed to the learning from a fusion of labeled data and the density distribution estimated from the unlabeled data, when maximizing the margin of separation.

## *B. Results on Real World Sentiment Datasets*

Here, we apply the proposed TOR on a real world application, particularly, Sentiment ordinal classification datasets.

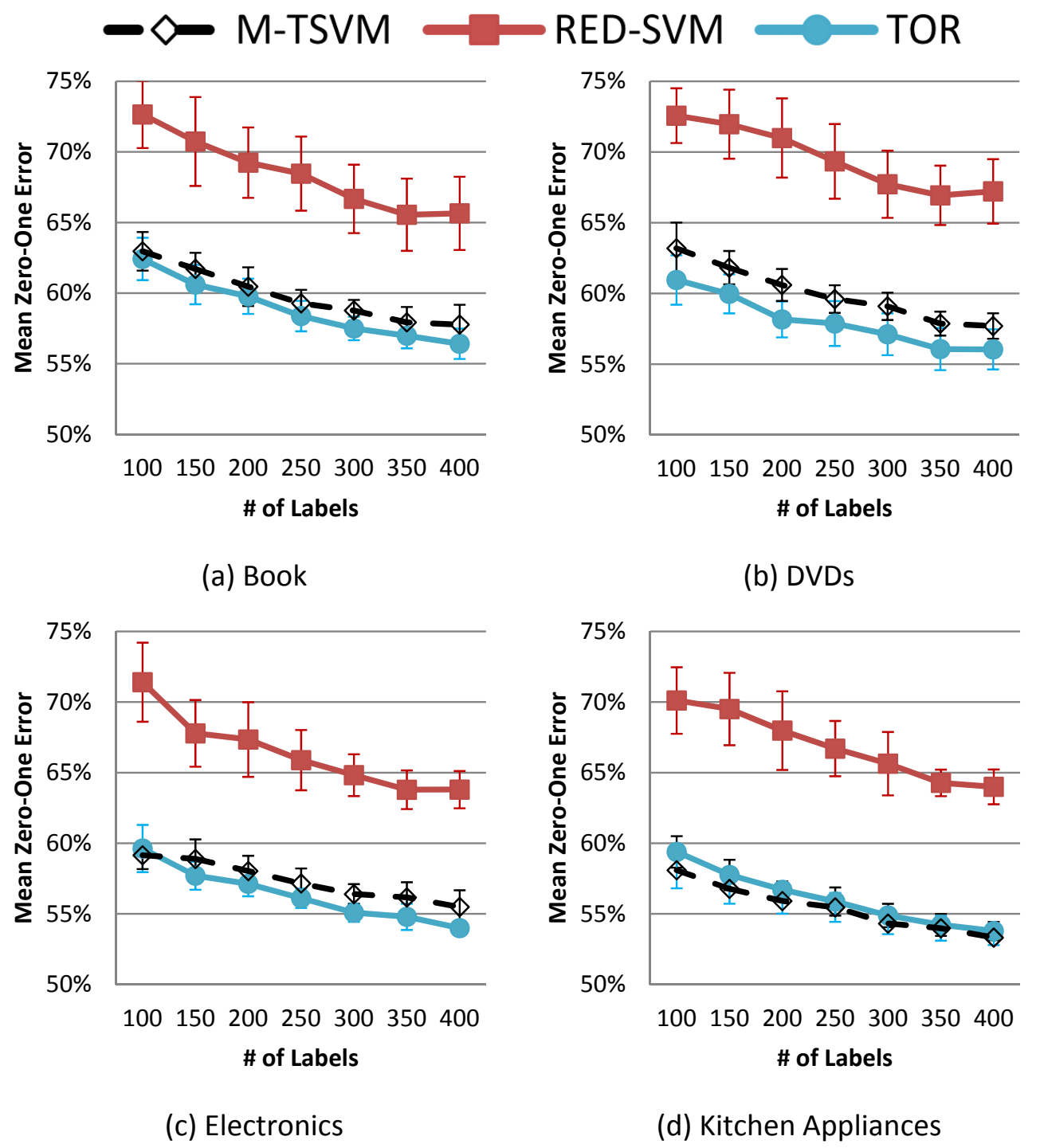

Fig. 10. Mean Zero-One Error on Sentiment datasets. Error bars denote the standard deviation

Since SVOR-EXC and SVOR-IMC are not designed to handle the datasets with inputs that are of high dimensions like the sentiment datasets, these two algorithms are omitted from the experimental study. The results obtained on the remaining algorithms are then summarized in Fig. 10.

Notably, TOR displayed superior performance over RED-SVM, with at least 8% and up to 12% improvements in accuracy. Furthermore, even though TOR employs only a small number of 100 labeled data samples, complimented by the unlabeled data, a significantly lower error relative to RED-SVM can be observed, despite the latter using a larger labeled data samples of 400. This observation clearly demonstrates the effectiveness of using unlabeled data in ordinal regression.

The mean absolute error metric defined in (13) is also reported for the sentiment dataset, as summarized in Fig. 11. It is worth noting that a mean absolute error larger than one indicates the average rating obtained differs from the true label by more than one rating scale. For example, RED-SVM with a mean absolute error close to one on labeled data of 100 indicates that the predicted labels of most samples differ from their respective true class labels by one unit. On the other hand, TOR is observed in Fig. 11 to exhibit significantly lower mean absolute error than the RED-SVM, thus suggesting that the predictions made by TOR are closer to the true labels on most data samples. Overall, TOR reports significantly lower mean absolute error than M-TSVM on all the datasets considered.

Another interesting observation that can be derived from Fig. 11 pertaining to limited labeled data available. Particularly, M-TSVM is shown to deliver a lower mean absolute error than RED-SVM under the condition of limited labeled data, which is made possible by complimenting the learning process with the abundant of unlabeled data. As the number of available labeled data increases, the ordinal information learned by RED-SVM generally helps to lower the mean absolute errors as observed in Fig. 11. In contrast, TOR benefited through learning from both the ordinal knowledge and the density information of unlabeled data to arrive at the improvements in mean absolute error observed over RED-SVM and M-TSVM.

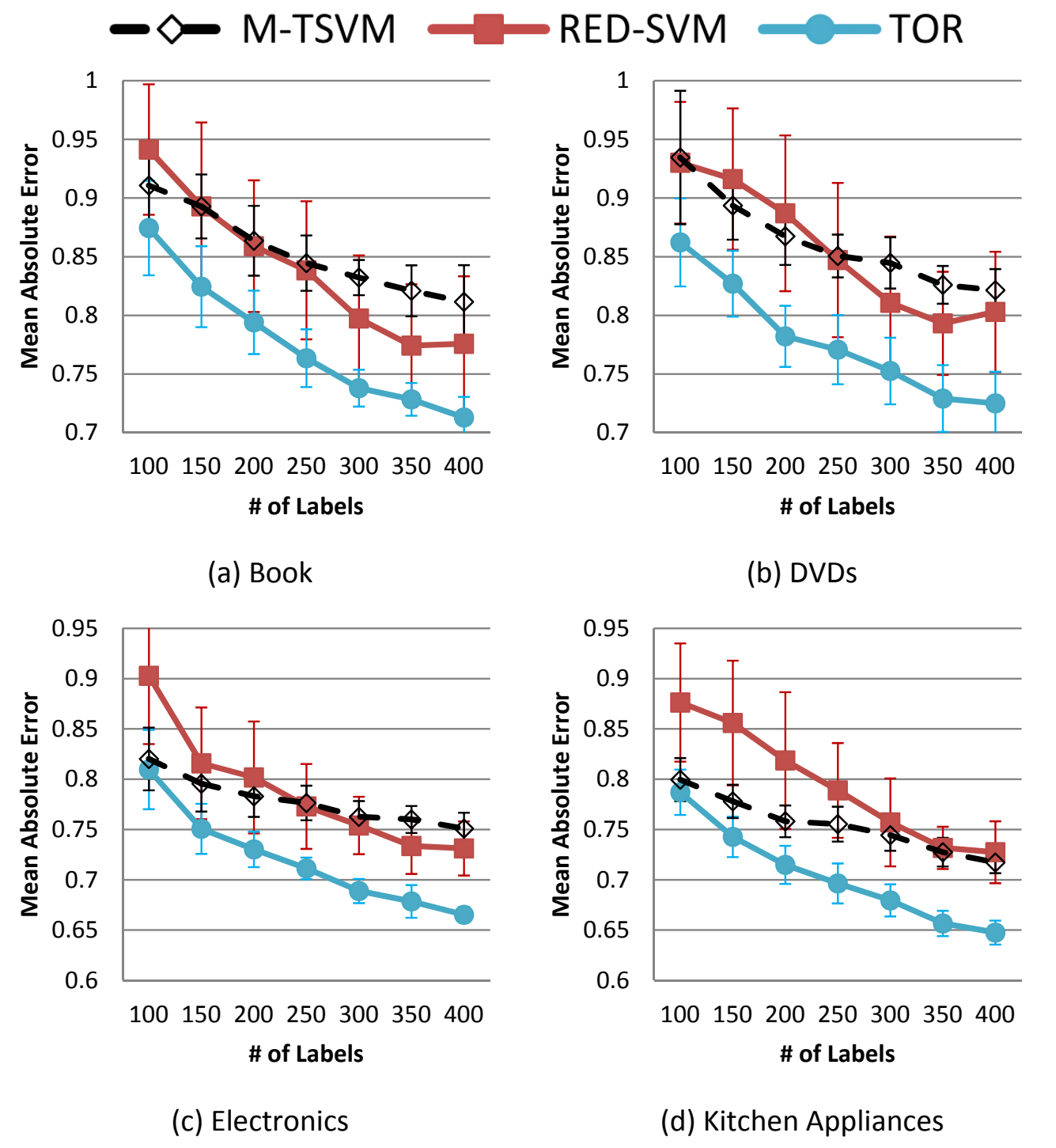

Fig. 11. Mean Absolute Error on Sentiment datasets. Error bars denote the standard deviation

In Figs. 10 and 11, the error bars representing the standard deviation are also presented[8]. As observed, the standard deviation obtained by the transductive algorithms, i.e., M-TSVM and TOR, are generally smaller than the inductive RED-SVM algorithm, thus acknowledging the robustness of the transductive learning paradigm.

Next, we analyze the label swapping procedure of the TOR in details by increasing the number of labels to be used to 2000. Fig. 12 depicts the effectiveness of label swapping after the label initialization. From the observations, label swapping effectively reduces the mean zero-one and absolute errors in Fig. 12(a) and 12(b), respectively, and while the number of labeled data increases, the improvements by TOR are decreasing. Another observation is that as the number of labeled data increases, the number of SVM training iterations within TOR will generally decrease as shown in Fig. 12(c). This is expected since as more labeled data are added into the training set, the decision boundaries become less affected by the unlabeled data. Therefore, the TOR is deem as more effective when only a small number of labeled data is available.

[8]For other figures on benchmark datasets, there are too many comparison algorithms depicted in those figures. Hence, the errors bars are not provided.

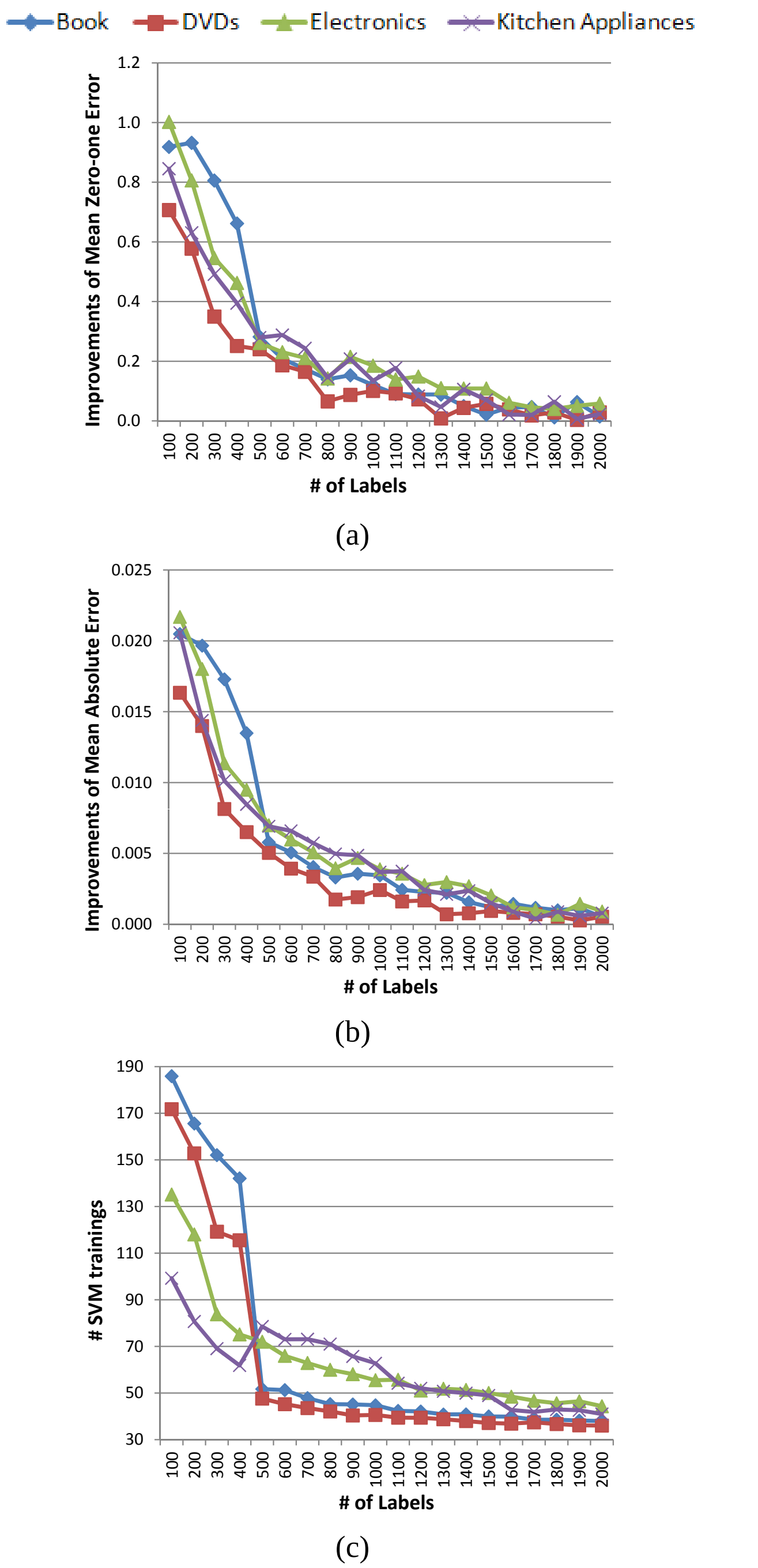

Fig. 12. Analysis of TOR after the label initialization phase. SubFigs. (a) and (b) depict the differences (improvements) of mean zero-one and mean absolute errors, respectively, between TOR reaches convergence in Algorithm 1 and after TOR initializes the labels. Subfig. (c) depicts the number of SVM trainings for TOR to reach convergence.

In Fig. 12(c), it depicts the number of iterations for TOR to converge. Let $T$ be the number of iterations for TOR to converge. The computational cost of TOR is then $O(TR)$, where $R$ be the computational cost of RED-SVM. However, it is notable here that the training process of TOR can be enhanced via a warm-start strategy, i.e., using the previous solution of the alpha variables as the initial alpha variables for the next iteration.

## VII. Conclusion

In this paper, by taking benefits from the abundance of unlabeled patterns, we had presented a novel transductive learning paradigm for ordinal regression, namely *Transductive Ordinal Regression* (TOR). To the best of our knowledge, the present work serves as the first attempt that addresses the general ordinal regression problem in a transductive setting for a family of ordinal loss functions. The family of ordinal loss functions including hinge loss, logistic loss and Laplacian loss are supported. A proposed label swapping scheme is also introduced to guarantee a strictly monotonic decrease in the objective value of the transductive ordinal function. Based on the experimental results obtained, TOR was reported to attain significant accuracy improvements over all the other algorithms considered via leveraging the cluster assumption on the unlabeled data and the ordinal constraints imposed to maximize the margin of separation between consecutive classes in ordinal regression. In situations where only few labeled data are available, TOR clearly serves as an indispensable tool.